\documentclass{article}

\PassOptionsToPackage{square,numbers}{natbib}

\usepackage[final]{neurips_2023}

\usepackage{mdframed}

\usepackage[utf8]{inputenc} 
\usepackage[T1]{fontenc}    
\usepackage{hyperref}       
\usepackage{url}            
\usepackage{booktabs}       
\usepackage{amsfonts}       
\usepackage{nicefrac}       
\usepackage{microtype}      
\usepackage{xcolor}         
\usepackage{graphicx,caption}
\usepackage{subcaption}

\usepackage{amsmath}
\usepackage{dirtytalk}
\usepackage{algorithm}
\usepackage{algpseudocode}
\usepackage{mathabx}
\usepackage{tikz}
\usetikzlibrary{positioning}
\usepackage{array, multirow, tabularx}
\usepackage{authblk}
\usepackage{ centernot}
\usepackage{amssymb}

\usetikzlibrary{arrows.meta, positioning}

\usepackage{scalerel} 
\newcommand{\CI}{\mathrel{\perp\mspace{-10mu}\perp}}
\newcommand{\nCI}{\mathrel{\ooalign{\hfil$\CI$\hfil\cr\hfil$\diagup$\hfil}}}

\graphicspath{ {./images/} }

\title{Spectral structure learning for clinical time series}

\author[1,2]{Ivan Lerner}
\author[1,2]{Anita Burgun}
\author[3]{Francis Bach}

\affil[1]{Université Paris Cité, Inria, Inserm, HeKA, F-75015 Paris, France}
\affil[2]{Service d'informatique biomédicale, AP-HP, Hôpital Georges Pompidou, F-75015 Paris, France}
\affil[3]{SIERRA, Inria Paris, F-75015 Paris, France}

\usepackage[acronym,xindy,toc]{glossaries} 

\makeglossaries

\newglossaryentry{latex}
{
        name=latex,
        description={Is a mark up language specially suited for 
scientific documents}
}

\newacronym{AP-HP}{AP-HP}{Assistance publique - Hôpitaux de Paris}
\newacronym{COVID-19}{COVID-19}{COronaVirus Disease 2019}
\newacronym{SARS-CoV-2}{SARS-CoV-2}{Severe Acute Respiratory Syndrome Coronavirus 2}
\newacronym{PharmWAS}{PharmWAS}{Pharmacopeia-Wide Association Study}
\newacronym{PheWAS}{PheWAS}{Phenome-Wide Association Study}
\newacronym{ICD-10}{ICD-10}{International Classification of Diseases, 10th edition}
\newacronym{CDW}{CDW}{Clinical Data Warehouse}
\newacronym{OMOP CDM}{OMOP CDM}{Observational Medical Outcomes Partnership's Common Data Model}
\newacronym{ATC}{ATC}{Anatomical Therapeutic Chemical}
\newacronym{NSAIDS}{NSAIDS}{Nonsteroidal anti-inflammatory drugs }
\newacronym{ACE}{ACE}{Angiotensin-Converting Enzyme}
\newacronym{ARB}{ARB}{Angiotensin Receptor Blockers}
\newacronym{PS}{PS}{Propensity Score}
\newacronym{BMI}{BMI}{Body Mass Index}

\newacronym{IPW}{IPW}{Inverse Probability Weighting}
\newacronym{EER}{EER}{Empirical Equipoise Region}

\newacronym{FBC}{FBC}{Fraction of Balanced Covariates}
\newacronym{ASMD}{ASMD}{Absolute Standardized Mean Difference}
\newacronym{FEP}{FEP}{Fraction of Exposed Population}
\newacronym{FMI}{FMI}{Fraction of Missing Information}
\newacronym{FDR}{FDR}{False Discovery Rate}

\newacronym{CRP}{CRP}{C-Reactive Protein}

\newacronym{$K_D$}{$K_D$}{Dissociation Constant}
\newacronym{RR}{RR}{Relative Risk}
\newacronym{OR}{OR}{Odds Ratio}
\newacronym{EHRs}{EHRs}{Electronic Health Records}

\newacronym{ICU}{ICU}{Intensive Care Unit}

\newacronym{LASSO}{LASSO}{Least Absolute Shrinkage and Selection Operator}

\newacronym{CG}{CG}{Conjugate Gradient}

\newacronym{GPU}{GPU}{Graphics Processing Unit}
\newacronym{GP}{GP}{Gaussian process}
\newacronym{DGP}{DGP}{data-generating process}
\newacronym{LTI}{LTI}{linear time-invariant}

\newacronym{MVN}{MVN}{multivariate normal}
\newacronym{p.s.d.}{\textit{p.s.d.}}{positive-semi definite}
\newacronym{nmll}{\textit{nmll}}{negative marginal log-likelihood}
\newacronym{MIMIC}{MIMIC}{Medical Information Mart for Intensive Care}
\newacronym{MBP}{MBP}{\textit{mean blood pressure}}

\newacronym{PC}{PC}{\textit{Peter \& Clarck}}
\newacronym{MCI}{MCI}{Momentary Conditional Independence}
\newacronym{PCMCI}{PC-MCI}{PC-Momentary Conditional Independence}

\newacronym{BIC}{BIC}{Bayesian information criterion}
\newacronym{DAG}{DAG}{directed acyclic graph}
\newacronym{SCM}{SCM}{structural causal model}
\newacronym{NOTEARS}{NOTEARS}{Non-combinatorial Optimization via Trace Exponential and Augmented lagRangian for Structure learning}
\newacronym{PGM}{PGM}{proximal gradient method}
\newacronym{SHD}{SHD}{structural hamming distance}
\newacronym{RMSE}{RMSE}{root mean square error}

\newacronym{SVAR}{SVAR}{structural vector autoregressive models}
\newacronym{VAR}{VAR}{vector autoregressive models}
\newacronym{DBN}{DBN}{dynamic bayesian network}
\newacronym{CSI}{CSI}{causal structure identification}
\newacronym{EM}{EM}{expectation-maximization}
\newacronym{StructGP}{StructGP}{Structured Gaussian process}
\newacronym{LPP-GP}{LPP-GP}{\textit{latent patient pathways Gaussian process}}

\newacronym{SSL}{SSL}{self-supervised learning }
\newacronym{NLP}{NLP}{Natural Langage Processing}

\newacronym{i.i.d.}{\textit{i.i.d.}}{\textit{independently and identically distributed}}
\newacronym{o.o.d.}{\textit{o.o.d.}}{\textit{out of distribution}}
\newacronym{p.d.f.}{\textit{p.d.f.}}{\textit{probability distribution function}}

\newacronym{SCD}{SCD}{sickle cell disease}
\newacronym{cEB}{cEB}{circulating erythroblasts}
\newacronym{ldh}{LDH}{Lactate Dehydrogenase}
\newacronym{hb}{Hb}{Hemoglobin}
\newacronym{hba2}{$HbA_2$}{Hemoglobin A2}
\newacronym{hbf}{HbF}{Fetal hemoglobin}
\newacronym{retics}{Retics}{Reticulocytes}
\newacronym{mcv}{MCV}{Mean Corpuscular Volume}
\newacronym{cr}{Cr}{creatinin}
\newacronym{wbc}{WBC}{white blood cells}
\newacronym{cb}{CB}{conjugate bilirubin}
\newacronym{hbs}{HbS}{Hemoglobin S}

\begin{document}

\maketitle

\begin{abstract}
We develop and evaluate a structure learning algorithm for clinical time series.
Clinical time series are multivariate time series observed in multiple patients and irregularly sampled, challenging existing structure learning algorithms.
We assume that our times series are realizations of \acrshort{StructGP}, a $k$-dimensional multi-output or multi-task stationary \acrfull{GP}, with independent patients sharing the same covariance function.
\acrshort{StructGP} encodes ordered conditional relations between time series, represented in a \acrlong{DAG}.
We implement an adapted NOTEARS algorithm, which based on a differentiable definition of acyclicity, recovers the graph by solving a series of continuous optimization problems.
Simulation results show that up to mean degree 3 and 20 tasks, we reach a median recall of 0.93\% [IQR, 0.86, 0.97] while keeping a median precision of 0.71\% [0.57-0.84], for recovering directed edges.
We further show that the regularization path is key to identifying the graph.
With \acrshort{StructGP}, we proposed a model of time series dependencies, that flexibly adapt to different time series regularity, while enabling us to learn these dependencies from observations.
\end{abstract}

\graphicspath{{images/}}

\section{Introduction}

Structure learning is the task of learning the dependency structure of either time-independent variables or, in this study, time series \cite{peters2017elements, glymour2019review}.
The structure of dependency is usually represented as a \acrfull{DAG}, in which nodes represent variables, and links between nodes represent relations between variables.
These links encode conditional or marginal independence relations \cite{koller2009probabilistic, lauritzen1996graphical}, or under certain strong additional assumptions (e.g., no hidden cofounders), have a causal interpretation and the structure can be used for \textit{causal modeling} \cite[Chapter 6]{peters2017elements}.

Structure learning algorithms are typically classified into constraint-based methods and score-based methods \cite{glymour2019review}.
Score-based methods assume a parametric model in which the parameters support the graph, and then search for the graph that best fits the data \cite{chickering2002optimal}.
The main limitation of this approach is that the acyclicity of the graph requires an iterative search over graph structures that satisfy the constraint.
And that the number of acyclic graphs grows superexponentially with the number of nodes in the graph \cite{robinson1977counting}.
Formulating the acyclicity of a \acrfull{DAG} as a differentiable function of its adjacency matrix allowed \citet{zheng2018dags} to frame the problem as a continuous optimization problem.
Following this line of work, DYNOTEARS \cite{pamfil2020dynotears} uses the acyclicity constraint to identify the structure of linear \acrfull{SVAR} \cite{demiralp2003searching}.

In this study, we aim to develop a structure learning algorithm for clinical time series collected from \acrfull{EHRs}.
\acrshort{EHRs} time series are collections of irregularly sampled multivariate time series, collected in multiple patients.
To learn graphical models of dependence between those, we work in continuous time and develop \acrshort{StructGP}, a structured multi-task \acrfull{GP} \cite{rasmussen2003gaussian}.

\section{Methods}

\subsection{Structured Gaussian process}\label{sec:structgp}

We consider the $k$-dimensional multi-output Gaussian process $\mathbf{Y}(t)$ defined as the  filtration by $\mathbf{H}(t)$ of white noise $\mathbf{w}(t)$:
\begin{align}\nonumber
    \mathbf{Y}(t) &= (\mathbf{H} * \mathbf{w})(t),
\end{align}
where $\mathbf{H}(t)$ is a sparse $k \times k$ lower triangular
matrix-valued impulse response function, and the components of
$\mathbf{w}(t)$ are independent unit-intensity Gaussian white noises.
An edge $u \to v$ means that the latent source $w_u$, associated with
series $Y_u$ through its self-filter, also contributes to $Y_v$
through $H_{vu}$. This is a variable ordering; it does not imply
temporal precedence or a causal effect of intervening on observed $Y_u$.

Taking the Fourier transform of $\textbf{Y}(t)$, we find $\textbf{Z}(\omega)$ a non-stationary complex white noise multi-output process \cite[p. 418]{papoulis2002probability}:
\begin{align}\label{eq:ZHW}
    &\mathbf{Z}(\omega) = \Tilde{\mathbf{H}}(\omega)  \mathbf{W}(\omega),
\end{align}
where $\Tilde{\mathbf{H}}(\omega)$ is the Fourier transform of $\mathbf{H}(t)$, and $\mathbf{W}(\omega)$ complex independent white noise processes.
In addition, the covariance of $\mathbf{Z}(\omega)$ or \textit{spectral density} is:
\[
\operatorname{Cov}\big( \mathbf{Z}(\omega), \mathbf{Z}(\omega') \big) = 
\begin{cases}
\Tilde{\mathbf{H}}(\omega) \Tilde{\mathbf{H}}^T(\omega) & \text{if } \omega = \omega' \\
0 & \text{if } \omega \neq \omega'
\end{cases}.
\]
Thus, it identify $\Tilde{\mathbf{H}}(\omega)$ as the Cholesky factor of the covariance matrix of $\mathbf{Z}(\omega)$.

The finite-dimensional identity in Equation~\ref{proof:oci},
Appendix~\ref{sec:proof:choloci}, has a spectral analogue.
For $u<v$, let $C=\{1,\ldots,u-1\}$. Then
\begin{align*}
    & \mathbf{H}_{vu}(t)=0,
      \quad \forall t\in\mathbb{R} \\
    \Leftrightarrow\quad
    & \widetilde{\mathbf{H}}_{vu}(\omega)=0,
      \quad \forall \omega\in\mathbb{R} \\
    \Leftrightarrow\quad
    & \mathbf{f}_{uv\mid C}(\omega)
      =\mathbf{f}_{uv}(\omega)
       -\mathbf{f}_{uC}(\omega)\mathbf{f}_{CC}^{-1}(\omega)
        \mathbf{f}_{Cv}(\omega)
      =0,
      \quad \forall \omega\in\mathbb{R} \\
    \Leftrightarrow\quad
    & Y_u \CI Y_v \mid \mathbf{Y}_C.
\end{align*}
Here, $\mathbf{f}_{uv\mid C}$ is the partial cross-spectrum,
and $Y_j$ denotes the complete latent trajectory.
For $C=\emptyset$, the partial cross-spectrum is simply
$\mathbf{f}_{uv}$.
The equivalence assumes spectral nondegeneracy and recovery of
preceding sources from complete preceding trajectories, as provided
by the triangular Gaussian-filter model.
It does not generally hold when conditioning only on finitely many
noisy observations.

We therefore parameterize $\mathbf{H}(t)$ as follows:
\begin{align}
    & \quad  \mathbf{H}(t) = (\mathbf{I} - \mathbf{S}) \circ \mathbf{L}(t), 
\end{align}
where  $\mathbf{L}_{vu}(t) = \exp\big(-\frac{t^2}{\mathbf{L}_{vu}}\big)$, $\mathbf{S}$  is a sparse lower triangular matrix up to permutation, $\mathbf{L}$ is a positive square matrix, $\mathbf{I}$ is the identity matrix, and $\circ$ the Hadamard product.
The support of $\textbf{S}$ can then be interpreted as the adjacency matrix of a \acrfull{DAG}  $\mathcal{G}$ that encodes ordered conditional independence relation:
\begin{align}
    \quad & \mathbf{S}_{vu} \neq 0 \\\nonumber
    \Leftrightarrow & Y_u \xrightarrow{\mathcal{G}} Y_v\\\nonumber
    \Leftrightarrow & Y_u \nCI Y_v \mid \mathbf{Y}_{\{1, 2, \ldots, u-1\}}.
\end{align}

These ordered conditional independences do not generally imply
a parent-only Markov factorization on $\mathcal{G}$.
Under the complete-trajectory assumptions above, a conventional
Markov factorization is available on its transitive closure,
with each variable conditioned on its ancestors in $\mathcal{G}$.

\subsection{Learning the graph}\label{sec:annealednotears}

As our time series are irregularly sampled from multiple patients we switch here to a set-based indexing, thanks to the marginalization properties of \acrshort{GP}.
Our data is then a collection of $n$ scalar observations $y$ from $r$ individuals and $k$ tasks,  $\{y(\mathbf{x}): \mathbf{x}\in (\mathbb{N}, \mathbb{N}, \mathbb{R}^+)\}$.
Each observation is indexed by the input vector $\mathbf{x}$, a triplet composed of the patient index $i$, task index $j$, and time $t$, such that $\mathbf{x} = (i, j, t)$, with $i \in \{1, ..., r\}$, $j \in \{1, ..., k\}$ and $t \in \mathbf{R}^+ $.
We observe multiple independent patients, and the intra-patient covariance of the process for patient $i$ can be written:
$$ \mathrm{Cov}[\mathbf{Y}(i, u, t) , \mathbf{Y}(i, v, t')] = (\mathbf{h}_u * \mathbf{h}_v^T)(t-t').$$
With classical \acrshort{GP}, the set of free parameters $\theta = \{\textbf{S}, \textbf{L} \}$  is learned by maximizing  $\log p(\mathbf{y}|\mathbf{X}, \theta)$, the marginal likelihood of the training observations $\mathbf{y}$ given inputs $\mathbf{X}$ \cite{mardia1984maximum}.
With structured \acrshort{GP}, we follow the \acrshort{NOTEARS} algorithm 
 to learn the graph, order of tasks, and sparsity pattern \cite{zheng2018dags} and impose an acyclicity constraint on  $\textbf{S}$.
\acrshort{NOTEARS} leverages the trace of the matrix exponential of the adjacency matrix as a differentiable acyclicity constraint.
Our objective is therefore to solve the constrained optimization problem below:
\begin{align}
  \theta^* &= \operatorname*{argmin}_\theta -\log p(\mathbf{y}, \mathbf{X}, \theta) + \lambda \| \textbf{S} \|_1 \\\nonumber
& \textit{s.t. }  \operatorname{Tr}(\exp(\mathbf{S} \circ \mathbf{S} )) - k = 0,
\end{align}
where $\lambda$ is the penalty strength.

The above is solved by dual ascent following the augmented Lagrangian method \cite{nemirovsky1999optimization}, such that the constrained problem is equivalent to solving a series of unconstrained problems, the primal and the dual.
The primal is the penalized objective function augmented with a quadratic penalty term, and is solved with a \acrlong{PGM} (see Appendix A\ref{sup:pgm}), the dual is solved by gradient ascent (see Appendix A\ref{sup:lagrang}).
Finally, we find by grid search $\lambda_*$, the sparsity penalty that minimizes an equivalent of the Akaike information criterion (AIC):
$\text{AIC} = 2\| \textbf{S} \|_0 - 2\log \mathcal{L}(\mathbf{y}, \mathbf{X}, \theta) .$
Grid search is conducted from $\lambda_{max}$ to $\lambda_{min}$ on a log-scale, with warm-start.
Because solving the augmented Lagrangian problem to small error is computationally expensive, and leads to numerical instability when $\rho$ becomes large, we choose to only loosely solve it, typically with a large tolerance for the acyclicity constraint ($\epsilon = 0.1$, see Appendix A\ref{sup:lagrang}).
Thus, to ensure dagness, we apply a hard-threshold operation, i.e. we mask elements in $\textbf{S}$ lower than the minimal threshold that ensures dagness.

\subsection{Simulation study}\label{sec:simstudy}
\begin{table}[h!]
\centering
\begin{tabularx}{\textwidth}{|c|X|X|X|X|}
  \hline
  \small
  \textbf{Experiment} & \textbf{Number of tasks} & \textbf{Mean \quad degree}  & \textbf{Grid search steps} & \textbf{Number of patients} \\
  & $k$ & $md$  & $n_{\lambda}$ & $r$\\
  \hline
  TOY & 4 & 2  & 256 & $  50$ \\
  EXP1 & 10 & 2  & 50 & $[1, \ldots, 100]$ \\
  EXP2 & 10 & 3  & $[2, \ldots, 512]$ & 50 \\
  EXP3 & $[2, \ldots, 20]$ & $[1, 2, 3]$  & 50 & 50 \\
  \hline
\end{tabularx}
\caption{Summary of simulation parameters}
\label{tab:struct_sim_param}
\end{table}
We empirically assess the accuracy of the algorithm to identify a graph from observations through a series of simulations.
For each simulation, we sample a graph, the covariance parameters $\theta$, and observations from the sampled prior \acrshort{GP}.
The definition of the \acrshort{GP} follows that of section \ref{sec:structgp}, with the additional constraint that lengthscales parameters are tied for each task and exponentiated (
$ \textbf{L}_{vu} = \exp(\ell_v) \text{ for all } v \in \{1, 2, \ldots, k\}, \; u \in \{1, 2, \ldots, k\} $).
We then fit the model using the overall algorithm from section \ref{sec:annealednotears} (including the grid search), and compare the predicted graph $\hat{\mathcal{G}}$ with the true simulated graph $\mathcal{G}$.
The comparison is made with the \acrfull{SHD}.
\acrshort{SHD} is the "edit distance" of graphs, it counts the number of edge modifications (insertion, deletion, inversion) necessary to transform a predicted graph into the true simulated graph.
We also compare the \acrfull{RMSE} between the predicted $\hat{\textbf{S}}$ and true (simulated) $\textbf{S}$ parameters.
Each simulation is repeated 100 times and we report the average metric along with its bootstrapped 95\% confidence intervals.
For comparison purposes, we also report the same metrics for a random graph from the same distribution as simulated.

In all simulations, the support of $\mathbf{S}$ is sampled from a random (Erd\H{o}s--R\'enyi) graph in which the sparsity level is controlled by the mean degree of the graph $md$.
$\mathbf{S}$ parameters are uniformly sampled in $[-2, -0.5] \cup [0.5, 2]$.
$\ell_u$ parameters are uniformly sampled in $[-0.5, 0.5]$.
The observation times, $t$, are uniformly sampled in $[0, 10]$.
The observation noise level is fixed at $\sigma = 0.01$ and given as oracle when learning.
We first report results from one simulation, 'TOY', a model with
4 tasks illustrating the graph interpretation and its recovery
from observations.
We then report 3 experiments each varying specific parameters (see Table \ref{tab:struct_sim_param}), while the number of observations per task is fixed at ($n_{k}=10$). 
Code available at \href{https://gitlab.com/lerner.ivan/spectral_sl.git}{\underline{gitlab}}.

\section{Results}

\subsection{Toy model}\label{sec:res:toy_graph}

\captionsetup{width=0.9\linewidth}

\begin{figure}[htbp!]
    \centering
    \begin{subfigure}[b]{0.3\textwidth}
        \centering
        \includegraphics[width=\textwidth]{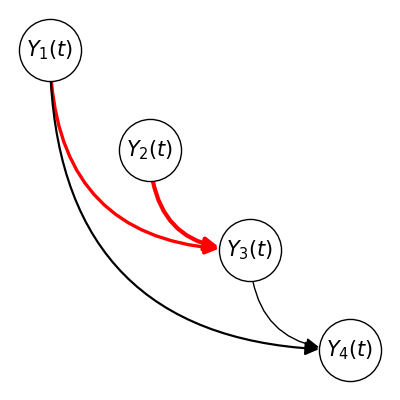}
        \vspace{0.7cm} 
        \caption{Simulated graph}
        \label{fig:toygraph}
    \end{subfigure}
    \hfill
    \begin{subfigure}[b]{0.6\textwidth}
        \centering
        \includegraphics[width=\textwidth]{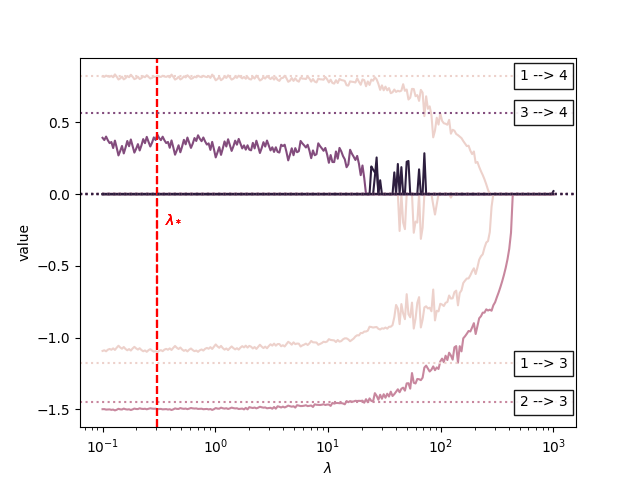}
        \caption{Regularization path}
        \label{fig:paths}
    \end{subfigure}
    \caption[A toy model with 4 tasks]{A toy model with 4 tasks \\
    \small
    We sample a random graph whose adjacency matrix sparsity pattern encodes ordered conditional relations between time series (a). 
    We recover the parameters and order of variables from observations sampled uniformly and independently between tasks (see section \ref{sec:simstudy}).
    }
    \label{fig:toypath}
\end{figure}

In Figure \ref{fig:toygraph}, we show a sampled random \acrshort{DAG} of mean degree 2 for 4 tasks, with 4 links.
A link $u \to v$ indicates a direct contribution of latent source
$w_u$ to output $Y_v$.
It corresponds to the following output scale parameters of the impulse response function $\textbf{H}(t)$:
\[
\textbf{I} - \textbf{S} = 
   \begin{bmatrix}
    1 & 0 & 0 & 0 \\
    0 & 1 & 0 & 0 \\
    -1.18 & -1.45 & 1 & 0 \\
    0.82 & 0 & 0.57 & 1 \\
    \end{bmatrix}.
\]
This graph encodes the following ordered independence relations:
\begin{align*}
     Y_4 \CI Y_2 &\mid Y_1,\\
     Y_2 \CI Y_1 &\mid \emptyset.
\end{align*}

Using the chain rule and $Y_2 \CI Y_1$, the complete-trajectory
law factorizes as follows, with products understood as products
of conditional probability kernels:
\begin{align*}
    P(\mathbf{Y})
    &= P(Y_4 \mid Y_3,Y_2,Y_1)
       P(Y_3 \mid Y_2,Y_1)
       P(Y_2 \mid Y_1)P(Y_1) \\
    &= P(Y_4 \mid Y_3,Y_2,Y_1)
       P(Y_3 \mid Y_2,Y_1)
       P(Y_2)P(Y_1).
\end{align*}
Although $Y_4 \CI Y_2 \mid Y_1$, conditioning additionally on
$Y_3$ generally induces dependence between $Y_4$ and $Y_2$.
Thus, $Y_2$ cannot be removed from the first factor.

Figure \ref{fig:paths} shows that with 50 patients and 10 observations per task, our algorithm exactly recovers the order of variables and the sparsity pattern.
The learned weights are slightly biased toward 0 due to the regularization properties of our objective function, the \acrshort{nmll}.

\subsection{Simulation study}\label{sec:res:graph_sim}
\begin{figure}[hbt!]
  \centering
  \includegraphics[width=0.9\textwidth]{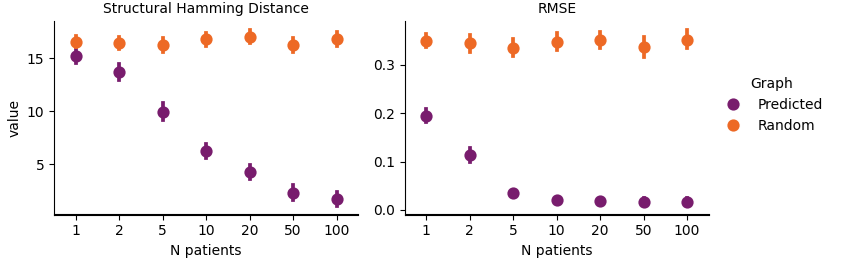}
  \captionsetup{width=0.9\linewidth}
  \caption[Average metrics for an increasing number of patients]{Average metrics for an increasing number of patients\\
  \small
  Reports of 'EXP1', increasing the number of patients for 10 tasks and 10 observations per task.
  The predicted graph (purple dots) is compared with a random graph from the same graph distribution (orange dots).
  The simulated graphs are random graphs with mean degree 2.
  Error bars represent bootstrap 95\% confidence intervals.
  }
  \label{fig:n_patients_shd}
\end{figure}
We report in Figure \ref{fig:n_patients_shd} the results of 'EXP1' which shows that the algorithm from section \ref{sec:annealednotears} identifies the true graph almost perfectly in this 'ideal' regime of parameters (large weights, sparse graphs, low noise).
Indeed, with 100 patients, most of the errors that contribute to the \acrshort{SHD} are extra links.
We then show in supplementary Figure S\ref{fig:n_lambda}, that the number of grid search steps is heavily impacting the capacity to recover the true graph (red dots), but not because it allows us to precisely infer the optimal $\lambda_*$ (green dots).
In fact, fitting the model from a random initialization with the optimal $\lambda_*$ produces very poor performances, and improving the number of grid search steps does not improve the solutions (green dots).
In supplementary Figure S\ref{fig:n_task_shd}, we see that up to mean degree 3 and 20 tasks, most errors are spurious links.
Table \ref{tab:sim_precision_recall} shows that, for recovering directed edges, we reach a median recall of 0.93\% [IQR, 0.86, 0.97] while keeping a median precision of 0.71\% [0.57-0.84] on $20$ nodes graphs.

\captionsetup{width=0.9\linewidth}
\begin{table}[htbp]
\centering
\resizebox{0.9\linewidth}{!}{%
\begin{tabular}{rrllll}
\toprule
  $k$ &  \textbf{P} (Predicted) &  \textbf{P} (Random)  & \textbf{R} (Predicted) &  \textbf{R} (Random)\\
\midrule
 4 &  1.00 [0.83-1.00] & 0.50 [0.33-0.67] & 1.00 [0.83-1.00] & 0.50 [0.33-0.67] \\
 10  & 0.82 [0.69-0.93] & 0.14 [0.08-0.25] & 0.94 [0.87-1.00] & 0.16 [0.07-0.25] \\
 20  & 0.71 [0.57-0.84] & 0.07 [0.04-0.11] & 0.93 [0.86-0.97] & 0.07 [0.04-0.12] \\
\bottomrule
\end{tabular}
}
\caption[Precision and recall]{Precision and recall \\
\small
Precision (\textbf{P}) and recall (\textbf{R}) median and interquartile for varying mean degree and number of task parameters, from 100 replications of 'EXP3' simulations with $md=3$.
False positives include extra links and reversed links.
Metrics are computed from models learned (Predicted) and compared with metrics for random graphs of the same distribution as the simulated data (Random).}
\label{tab:sim_precision_recall}
\end{table}

\section{Discussion}
We developed a structure learning algorithm that learns ordered conditional independence relations between irregularly sampled time series.
These relations are parametrized by \acrshort{StructGP}, a \acrshort{GP} model built upon the convolution between a sparse lower-triangular matrix-valued impulse response function and independent white noises.
It corresponds to assuming a linear additive Gaussian \acrshort{SCM} for the Fourier representation of the time series, whose structure is invariant over all frequencies.
Based on a differentiable definition of acyclicity, this algorithm recovers the true graph by solving a series of continuous optimization problems with high sensitivity and good precision on simulated data.

The recent work by \citet{dallakyan2023learning} is the closest to ours.
They develop a structure learning algorithm for time series by imposing a Gaussian linear additive \acrshort{SCM} on the discrete Fourier transform of the time series.
However, they learn different weight matrices at each frequency, whereas we assume an invariant structure across continuous frequencies, parametrized by only one weight matrix.

More work will be needed to bridge the gap between simulated data and real-world data with regard to the sensitivity to standardization and unmeasured cofounders.
Indeed, when standardizing the time series, it was reported that with time-independent variables, standardization affected graph recovery~\cite{kaiser2022unsuitability, reisach2021beware}.
However, it is possible to re-parameterise the \acrshort{SCM} to ensure a marginal unit variance without losing the identifiability of the graph from observations \cite{ormaniec2024standardizing}.
Furthermore, scaling to large datasets could be achieved, for instance, with GPU-friendly solvers that exploit block sparsity induced by independence between patients \cite{gardner2018gpytorch}.

\bibliography{reference.bib}

@article{nemirovsky1999optimization,
  title={Optimization II. Numerical methods for nonlinear continuous optimization},
  author={Nemirovsky, AS},
  year={1999},
  publisher={Technion--Israel Institute of Technology}
}

@article{gardner2018gpytorch,
  title={Gpytorch: Blackbox matrix-matrix gaussian process inference with GPU acceleration},
  author={Gardner, Jacob and Pleiss, Geoff and Weinberger, Kilian Q and Bindel, David and Wilson, Andrew G},
  journal={Advances in Neural Information Processing Systems},
  volume={31},
  year={2018}
}

@article{ormaniec2024standardizing,
  title={Standardizing Structural Causal Models},
  author={Ormaniec, Weronika and Sussex, Scott and Lorch, Lars and Sch{\"o}lkopf, Bernhard and Krause, Andreas},
  journal={arXiv preprint arXiv:2406.11601},
  year={2024}
}

@article{reisach2021beware,
  title={Beware of the simulated dag! causal discovery benchmarks may be easy to game},
  author={Reisach, Alexander and Seiler, Christof and Weichwald, Sebastian},
  journal={Advances in Neural Information Processing Systems},
  volume={34},
  pages={27772--27784},
  year={2021}
}

@article{kaiser2022unsuitability,
  title={Unsuitability of NOTEARS for causal graph discovery when dealing with dimensional quantities},
  author={Kaiser, Marcus and Sipos, Maksim},
  journal={Neural Processing Letters},
  volume={54},
  number={3},
  pages={1587--1595},
  year={2022},
  publisher={Springer}
}

@book{papoulis2002probability,
  title={Probability, Random Variables and Stochastic Processes by Athanasios},
  author={Papoulis, S},
  year={2002},
  publisher={Boston: McGraw-Hill}
}

@article{jurek2022hierarchical,
  title={Hierarchical sparse Cholesky decomposition with applications to high-dimensional spatio-temporal filtering},
  author={Jurek, Marcin and Katzfuss, Matthias},
  journal={Statistics and Computing},
  volume={32},
  number={1},
  pages={15},
  year={2022},
  publisher={Springer}
}

@inproceedings{robinson1977counting,
  title={Counting unlabeled acyclic digraphs},
  author={Robinson, Robert W},
  booktitle={Combinatorial Mathematics V: Proceedings of the Fifth Australian Conference, Held at the Royal Melbourne Institute of Technology, August 24--26, 1976},
  pages={28--43},
  year={1977},
  organization={Springer}
}

@book{koller2009probabilistic,
  title={Probabilistic graphical models: principles and techniques},
  author={Koller, Daphne and Friedman, Nir},
  year={2009},
  publisher={MIT press}
}

@book{lauritzen1996graphical,
  title={Graphical models},
  author={Lauritzen, Steffen L},
  volume={17},
  year={1996},
  publisher={Clarendon Press}
}

@article{beck2009gradient,
  title={Gradient-based algorithms with applications to signal recovery},
  author={Beck, Amir and Teboulle, Marc},
  journal={Convex optimization in signal processing and communications},
  pages={42--88},
  year={2009}
}

@article{parikh2014proximal,
  title={Proximal algorithms},
  author={Parikh, Neal and Boyd, Stephen},
  journal={Foundations and Trends in optimization},
  volume={1},
  number={3},
  pages={127--239},
  year={2014},
  publisher={Now Publishers Inc. Hanover, MA, USA}
}

@book{peters2017elements,
  title={Elements of causal inference: foundations and learning algorithms},
  author={Peters, Jonas and Janzing, Dominik and Sch{\"o}lkopf, Bernhard},
  year={2017},
  publisher={The MIT Press}
}

@article{dallakyan2023learning,
  title={On Learning Time Series Summary DAGs: A Frequency Domain Approach},
  author={Dallakyan, Aramayis},
  journal={arXiv preprint arXiv:2304.08482},
  year={2023}
}

@article{glymour2019review,
  title={Review of causal discovery methods based on graphical models},
  author={Glymour, Clark and Zhang, Kun and Spirtes, Peter},
  journal={Frontiers in genetics},
  volume={10},
  pages={524},
  year={2019},
  publisher={Frontiers}
}

@inproceedings{pamfil2020dynotears,
  title={Dynotears: Structure learning from time-series data},
  author={Pamfil, Roxana and Sriwattanaworachai, Nisara and Desai, Shaan and Pilgerstorfer, Philip and Georgatzis, Konstantinos and Beaumont, Paul and Aragam, Bryon},
  booktitle={International Conference on Artificial Intelligence and Statistics},
  pages={1595--1605},
  year={2020},
  organization={PMLR}
}

@article{demiralp2003searching,
  title={Searching for the causal structure of a vector autoregression},
  author={Demiralp, Selva and Hoover, Kevin D},
  journal={Oxford Bulletin of Economics and statistics},
  volume={65},
  pages={745--767},
  year={2003},
  publisher={Wiley Online Library}
}

@article{chickering2002optimal,
  title={Optimal structure identification with greedy search},
  author={Chickering, David Maxwell},
  journal={Journal of machine learning research},
  volume={3},
  number={Nov},
  pages={507--554},
  year={2002}
}

@article{zheng2018dags,
  title={Dags with no tears: Continuous optimization for structure learning},
  author={Zheng, Xun and Aragam, Bryon and Ravikumar, Pradeep and Xing, Eric P},
  journal={arXiv preprint arXiv:1803.01422},
  year={2018}
}

@article{mardia1984maximum,
  title={Maximum likelihood estimation of models for residual covariance in spatial regression},
  author={Mardia, Kanti V and Marshall, Roger J},
  journal={Biometrika},
  volume={71},
  number={1},
  pages={135--146},
  year={1984},
  publisher={Oxford University Press}
}

@incollection{rasmussen2003gaussian,
  title={Gaussian processes in machine learning},
  author={Rasmussen, Carl Edward},
  booktitle={Summer school on machine learning},
  pages={63--71},
  year={2003},
  publisher={Springer}
}

\section{List of abbreviations}

\printglossary[type=\acronymtype, title=]

\section{Appendix} 

\subsection{Conditional independence}\label{sec:proof:ci}

These sections adapt proofs from \cite{lauritzen1996graphical, jurek2022hierarchical} with the same notations as the remainder of the article for simplicity.

\subsection{Sparsity pattern of the precision matrix encodes conditional independence}\label{sec:proof:precisionci}

Let \(\mathbf{Y} = (Y_1, Y_2, \ldots, Y_k)^\top\) be a multivariate normal random vector with mean vector \(\mu\) and covariance matrix \(K\), i.e., \(\mathbf{Y} \sim \mathcal{N}(\mu, K)\).

Consider the two sets of variables \(A = \{u, v\}\) and \(B = \{1, 2, \ldots, k\} \setminus \{u, v\}\). The covariance matrix can be partitioned as:
$$ 
K =
     \begin{pmatrix}
        \textbf{K}_A & \textbf{K}_{AB} \\
        \textbf{K}_{AB}^T &  \textbf{K}_{B}
    \end{pmatrix}.
$$

After conditioning on \(B\), the conditional covariance matrix is given by:
$$ 
\textbf{K}_{A \mid B} = \textbf{K}_A - \textbf{K}_{AB}\textbf{K}_B^{-1}\textbf{K}_{AB}^T.
$$

Now, define the precision matrix \(\mathbf{\Omega} = \textbf{K}^{-1}\) with the same partition:
$$
\mathbf{\Omega} = 
\begin{pmatrix}
    \mathbf{\Omega}_A & \mathbf{\Omega}_{AB} \\
    \mathbf{\Omega}_{AB}^T &  \mathbf{\Omega}_{B}
\end{pmatrix}.
$$

Using the inversion of a partitioned matrix, we recognize that:
$$
\mathbf{\Omega}_A = (\textbf{K}_A - \textbf{K}_{AB}\textbf{K}_B^{-1}\textbf{K}_{AB}^T )^{-1}.
$$

Or otherwise stated that the precision matrix is invariant to conditioning:
$$
\mathbf{\Omega}_{A \mid B} = \textbf{K}_{A \mid B}^{-1} = \mathbf{\Omega}_{A}.
$$

Thus, let:
$$
\mathbf{\Omega}_{A} = \begin{pmatrix}
\omega_{uu} & \omega_{uv}\\
\omega_{uv} & \omega_{vv}\\
\end{pmatrix}.
$$

Then, we have:
$$
\textbf{K}_{A \mid B} = \frac{1}{\det(\mathbf{\Omega}_{A})} \begin{pmatrix}
\omega_{vv} & -\omega_{uv} \\
-\omega_{uv} & \omega_{uu}
\end{pmatrix}.
$$

This shows that:
$$
Y_u \CI Y_v \mid \mathbf{Y}_{\setminus \{u,v\}} \Leftrightarrow \omega_{uv} = 0.
$$

\subsection{Sparsity pattern of the Cholesky factor of the covariance matrix encodes ordered conditional independence}\label{sec:proof:choloci}

Let \(\mathbf{Y} = (Y_1, Y_2, \ldots, Y_k)^\top\) be a multivariate normal random vector with mean vector \(\mu\) and covariance matrix \(\textbf{K}\), i.e., \(\mathbf{Y} \sim \mathcal{N}(\mu, \textbf{K})\).

Consider the two sets of variables $A = \{1, 2, \ldots, u-1\}$ and $B = \{u, \ldots, k\}$. The covariance matrix $\textbf{K}$ can be partitioned and factorized as follows:
\begin{align*}\\
    \textbf{K} &=
     \begin{pmatrix}
        \textbf{K}_A & \textbf{K}_{AB} \\
        \textbf{K}_{AB}^T &  \textbf{K}_{B}
    \end{pmatrix} \\
    &= \begin{pmatrix}
        I & 0 \\
        \textbf{K}_{AB}^\top \textbf{K}_A^{-1} & I
        \end{pmatrix}
        \begin{pmatrix}
        \textbf{K}_A & 0 \\
        0 & \textbf{K}_B - \textbf{K}_{AB}^\top \textbf{K}_A^{-1} \textbf{K}_{AB}
        \end{pmatrix}
        \begin{pmatrix}
        I & \textbf{K}_A^{-1} \textbf{K}_{AB} \\
        0 & I
    \end{pmatrix}.
\end{align*}

Hence, the Cholesky factor $\textbf{L}$ of $\textbf{K}$ is given by:
\begin{align*}
\textbf{L} &= 
    \begin{pmatrix}
        \textbf{L}_A & 0 \\
        \textbf{K}_{AB}^\top \textbf{L}_A^{-1} & \textbf{L}_S
        \end{pmatrix}, \\
        \text{where } \\
        \textbf{L}_A &= \text{chol}(\textbf{K}_A), \\\nonumber \text{ and } \textbf{L}_S &= \text{chol}(\textbf{K}_B - \textbf{K}_{AB}^\top \textbf{K}_A^{-1} \textbf{K}_{AB}).
\end{align*}

The element of the $v$-th row and $u$-th column, $\textbf{L}_{vu}$, can be found in $\mathbf{L}_S$ in the first column of at the row $(v-u+1)$:
$$\textbf{L}_{vu} = \textbf{L}_{S, (v-u+1)1}.$$

The conditional covariance given $A$ (i.e., all the variables preceding $u$) is:
$$\textbf{K}_{B \mid A} = \mathbf{L}_S \mathbf{L}_S^T.$$

This shows that:
\begin{align}
    & Y_u \CI Y_v \mid \mathbf{Y}_{\{1, 2, \ldots, u-1\}} \label{proof:oci}\\\nonumber
    \Leftrightarrow \quad & \textbf{K}_{B \mid A, v1} = 0 \\\nonumber
    \Leftrightarrow \quad & \sum_{u=1}^{k} \textbf{L}_{S,(v-u+1)u} \textbf{L}_{S, 1u} = 0\\\nonumber
    \Leftrightarrow \quad &  \textbf{L}_{S, (v-u+1)1} = 0\\\nonumber
    \Leftrightarrow \quad & \textbf{L}_{vu} = 0.
\end{align}

\newpage
\subsection{The augmented Lagrangian optimization algorithm}\label{sup:lagrang}

\begin{mdframed}[linewidth=1.5pt, 
    innerleftmargin=10pt, 
    innerrightmargin=10pt, 
    frametitle={}]
Given:
\begin{itemize}
    \item Objective function: $f(\theta) = -\log \mathcal{L}(\mathbf{y}, \mathbf{X}, \theta) + \mathcal{P}_{\lambda}(\mathbf{S})$
    \item Constraint: $g(\theta) = \mathbf{tr}(\exp(\mathbf{S} \circ \mathbf{S} )) - k = 0$
    \item Convergence criteria: constraint tolerance $\epsilon$ and $\rho_{max}$
    \item Primal solver: \textit{Solver}
\end{itemize}
Define:
\begin{itemize}
    \item Lagrange multiplier $\alpha^{(k)}$ at step $k$
    \item Augmented Lagrangian: $L(\theta, \alpha^{(k)}, \rho) = f(\theta) + \alpha^{(k)} g(\theta) + \frac{\rho}{2}g(\theta)^2$
\end{itemize}
Do:
\begin{enumerate}
    \item Choose initial guess $\theta^{(0)}$, Lagrange multipliers $\alpha^{(0)}=0$ and $\rho = 1$
    \item For $k = 0, 1, 2, \ldots$
    \begin{enumerate}
        \item  While $\rho < \rho_{max}$ update $\theta^{(k+1)}$.
        \begin{itemize}
            \item  Minimize the augmented Lagrangian function:
            \[
            \theta^{(k+1)} = \textit{Solver}\big(L(\theta, \alpha^{(k)}, \rho)\big)
            \]
            \item Break if $g(.)$ sufficiently decreases:
            \[g(\theta^{(k+1)}) < 0.25\ g(\theta^{(k)}) \]
            \item Else augment $\rho$: 
            \[\rho = 10 \ \rho \]
        \end{itemize}
        \item Update Lagrange multipliers:
        \[
        \alpha^{(k+1)} = \alpha^{(k)} + \rho \cdot g(\theta^{(k+1)})
        \]
        \item Check convergence criteria. If satisfied, stop.
        \[
         g(\theta^{(k+1)}) < \epsilon \text{ or } \rho \geq \rho_{max}
        \]
    \end{enumerate}
\end{enumerate}
\end{mdframed}

\subsection{Proximal gradient method}\label{sup:pgm}

\acrshort{PGM} \cite{parikh2014proximal} is a generalization of gradient descent for objective functions that can be split between a differentiable and non-differentiable part.
It requires defining a proximal operator for the non-differentiable part:
\begin{flalign}
\textbf{prox}_{\lambda}(\mathbf{S})_{uv} &= \begin{cases}
s_{uv} - \lambda, &\text{ if } s_{uv} > \lambda\\\nonumber
0, &\text{ if } s_{uv} \leq |\lambda | \\\nonumber
s_{uv} + \lambda, &\text{ if } s_{uv} <  -\lambda.
\end{cases}
\end{flalign}
Parameters are found by taking a classic gradient step followed by a proximal step at each iteration $k$:
\begin{flalign}
\text{(i) } \theta^{k+\frac{1}{2}} &= \theta^k - \alpha^k \nabla (- \log \mathcal{L}(\mathbf{y},  \mathbf{X}, \theta^k)) \nonumber \\
\text{(ii) } \theta^{k+1} &= \textbf{prox}_{\alpha^k\lambda} (\theta^{k+\frac{1}{2} }) \nonumber
\end{flalign}
with $\alpha^k$ a learning rate determined by line search at each iteration, following Beck and Teboulle \cite{beck2009gradient}.

\newpage
\subsection{Simulation study additional results}

\begin{figure}[hbt!]
  \centering
  \includegraphics[width=0.9\textwidth]{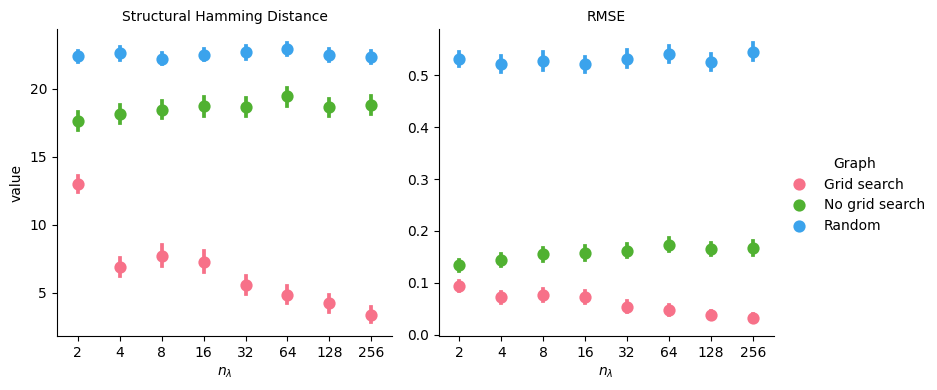}
  \captionsetup{width=0.9\linewidth}
  \caption[Average metrics for an increasing number of grid search steps ($n_{\lambda}$)]{Average metrics for an increasing number of grid search steps ($n_{\lambda}$)\\
  \small
  Reports of 'EXP2', an increasing number of grid search steps for 50 patients, 10 tasks, 10 observations per task, and random graph of mean degree 3.
  The predicted graph (red dots) is compared with a random graph from the same graph distribution (blue dots), and also with a predicted graph directly fitted from a random initialization with the optimal $\lambda_*$ found from grid search (green dots).
  Error bars represent bootstrap 95\% confidence intervals.
  }
  \label{fig:n_lambda}
\end{figure}
\newpage

\begin{figure}[h!]
  \centering
  \includegraphics[scale=0.5]{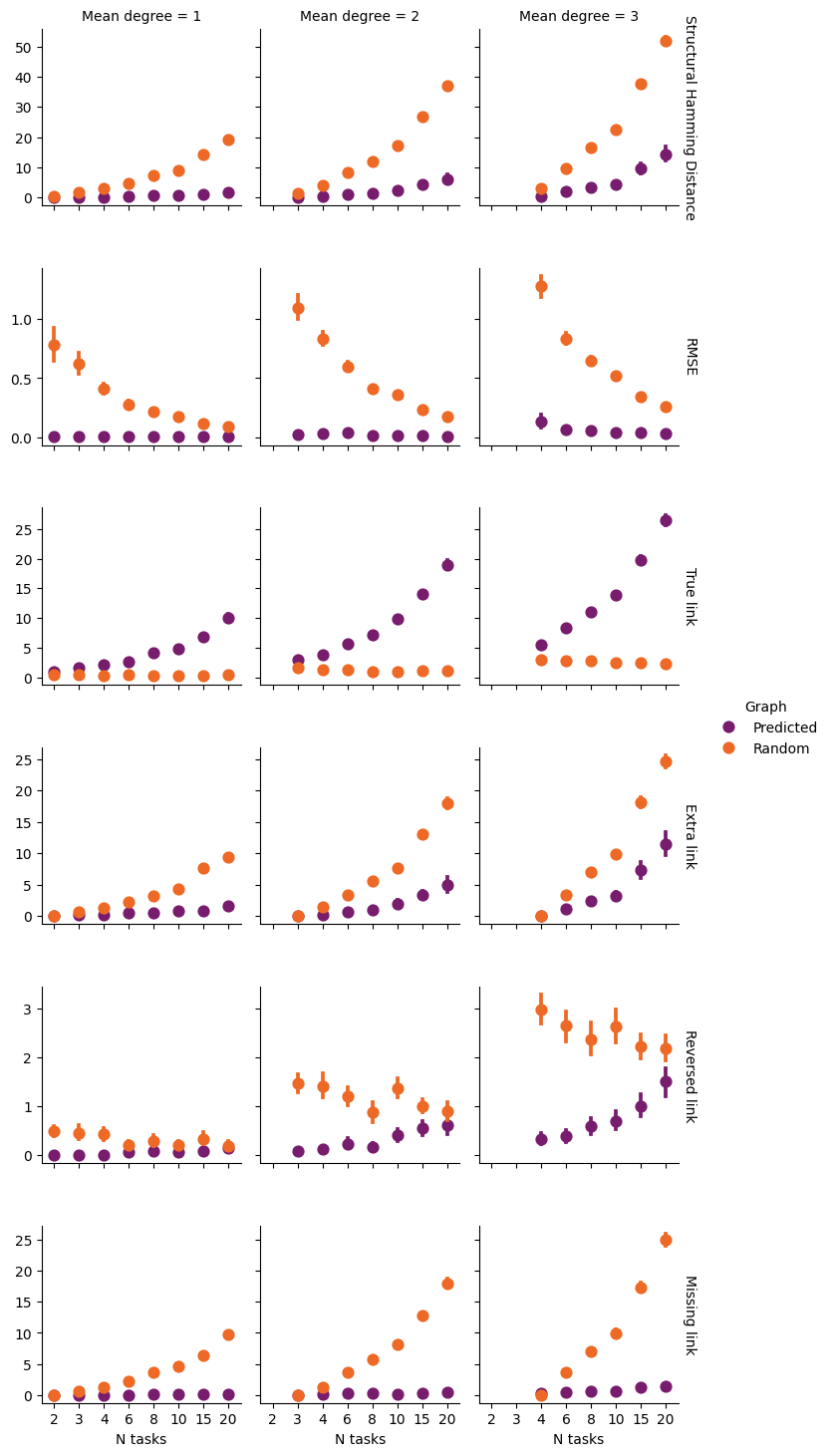}
  \captionsetup{width=0.9\linewidth}
  \caption[Average metrics for an increasing number of tasks]{Average metrics for an increasing number of tasks\\
  \small
  Reports of 'EXP3', an increasing number of tasks for 50 patients, and 10 observations per task.
  The predicted graph (purple dots) is compared with a random graph from the same graph distribution (orange dots).
  Error bars represent bootstrap 95\% confidence intervals.
  }
  \label{fig:n_task_shd}
\end{figure}


\end{document}